%% file: paper.tex
\documentclass[runningheads]{llncs}
\usepackage[T1]{fontenc}

\input{custom_packages.tex}
\input{custom_definitions.tex}
\input{acronyms.tex}

\newcommand\citet[1]{\cite{#1}}
\newenvironment{thm}{\begin{theorem}}{\end{theorem}}

\usepackage[hidelinks]{hyperref} 
\usepackage[capitalize,noabbrev]{cleveref}  
\crefformat{equation}{(#2#1#3)}

\begin{document}

\title{Exploiting Chordal Sparsity for Fast Global Optimality with Application to Localization}
\titlerunning{Exploiting Chordal Sparsity for Fast Global Optimality}
%
\author{Frederike Dümbgen\inst{1}\orcidID{0000-0002-7258-9753} \and
Connor Holmes\inst{2}\orcidID{0000-0002-8314-3677} \and
Timothy D. Barfoot\inst{2}\orcidID{0000-0003-3899-631X}}
\authorrunning{F. Dümbgen et al.}
%
\institute{
Inria, Ecole Normale Supérieure, PSL University, Paris, France\\
\url{https://di.ens.fr/willow/}\\
\email{frederike.dumbgen@inria.fr} \and
University of Toronto, Robotics Institute, Toronto ON M3H5T6, Canada\\
\url{https://robotics.utoronto.ca}}

\maketitle              

\begin{abstract}

\include*{_sections/abstract}

\end{abstract}
%

\acresetall

\include*{_sections/intro}

\include*{_sections/related_work}

\include*{_sections/background}
\include*{_sections/method}
\include*{_sections/algos}
\include*{_sections/results1}
\include*{_sections/conclusion}

\begin{credits}
\subsubsection{\ackname} 

We would like to thank Matt Giamou (McMaster) and Alexandre Amice (MIT) for introducing us to the idea of exploiting chordal sparsity in SDPs. FD is funded by the ANR JCJC project NIMBLE (ANR-22-CE33-0008), and most of this work was conducted while at the University of Toronto, funded by the Swiss National Science Foundation, Postdoc Mobility under Grant 206954. The Natural Sciences and Engineering Research Council of Canada also partly funded this work. 

\subsubsection{\discintname}

The authors have no competing interests to declare that are
relevant to the content of this article. 

\end{credits}
\include*{_sections/appendix}

\bibliographystyle{splncs04}
\bibliography{zotero-final}

\end{document}

%% file: custom_packages.tex
\usepackage[utf8]{inputenc} 
\usepackage{graphicx}
\usepackage{amsfonts}
\usepackage{mathtools}
\usepackage{siunitx}
\usepackage{tabularx}
\usepackage[table]{xcolor}
\usepackage{booktabs}
\usepackage{bm}
\usepackage[nolist,nohyperlinks]{acronym}
\usepackage{newclude}
\usepackage[textsize=tiny,color=gray!20]{todonotes}

\usepackage{tikz}
\usetikzlibrary{matrix}

%% file: custom_definitions.tex
\newcommand{\vc}[1]{\ensuremath{\bm{#1}}}

\newcommand{\bmat}[1]{\ensuremath{\begin{bmatrix}#1\end{bmatrix}}}

\DeclarePairedDelimiter\abs{\lvert}{\rvert}
\DeclarePairedDelimiter\norm{\lVert}{\rVert}
\makeatletter
\let\oldabs\abs
\let\oldnorm\norm
\def\abs{\@ifstar{\oldabs}{\oldabs*}}
\def\norm{\@ifstar{\oldnorm}{\oldnorm*}}
\makeatother

\DeclareMathOperator{\rank}{rank}
\DeclareMathOperator{\trace}{tr}
\DeclareMathOperator{\vech}{vech}
\DeclareMathOperator{\vechh}{vech}
\DeclareMathOperator{\mat}{mat}

\def\-{\raisebox{0pt}{-}}

\newcommand{\argmin}[1]{\underset{#1}{\operatorname{arg}\,\operatorname{min}}\;}

\newcommand{\R}[1]{\ensuremath{\boldsymbol{\mathbb{R}}^{#1}}}
\newcommand{\inR}[1]{\ensuremath{\in \R{#1}}}

\newcommand{\SO}[1]{\ensuremath{SO({#1})}}
\newcommand{\SE}[1]{\ensuremath{SE({#1})}}

\newcommand{\tr}[1]{\trace{\left(#1\right)}}

\let\oldunderbrace\underbrace
\renewcommand{\underbrace}[2]{\let\scriptstyle\textstyle \oldunderbrace{#1}_{#2}}
\definecolor{tab10blue}{rgb}{0.12156863, 0.46666667, 0.70588235}

\def\eg{\emph{e.g}.} 
\def\ie{\emph{i.e}.}

\DeclareFontFamily{U}{mathx}{\hyphenchar\font45}
\DeclareFontShape{U}{mathx}{m}{n}{ <-> mathx10 }{}
\DeclareSymbolFont{mathx}{U}{mathx}{m}{n}
\DeclareFontSubstitution{U}{mathx}{m}{n}
\DeclareMathAccent{\widebar}{\mathalpha}{mathx}{"73}
\newcolumntype{C}[1]{>{\centering\let\newline\\\arraybackslash\hspace{0pt}}m{#1}}
\newcolumntype{D}[1]{>{\centering\let\newline\\\arraybackslash\hspace{0pt}\columncolor{gray!20}}m{#1}}
\newcolumntype{R}[1]{>{\raggedleft\let\newline\\\arraybackslash\hspace{0pt}}m{#1}}
\newcolumntype{L}[1]{>{\raggedright\let\newline\\\arraybackslash\hspace{0pt}}m{#1}}
\newcolumntype{M}[1]{>{\raggedright\let\newline\\\arraybackslash\hspace{0pt}}m{#1}}
\newcolumntype{N}{>{\refstepcounter{rowcntr}\therowcntr}c}

\usepackage{tabstackengine}
\stackMath
\setstackgap{L}{18pt}
\setstacktabbedgap{0pt}
\def\lrgap{\kern6pt}

  \renewenvironment{thebibliography}[1]{%
    \begin{oldthebibliography}{#1}%
      \setlength{\parskip}{0ex}%
      \setlength{\itemsep}{0ex}%
  }%
  {%
    \end{oldthebibliography}%
  }

\usepackage{pifont}

\newcommand{\exampleRO}{\ac{CTRO}}
\newcommand{\exampleMW}{\ac{MW}}
\newcommand{\exampleheadRO}{\vspace{-1em}\subsubsection{Example CT-RO.}}
\newcommand{\exampleheadMW}{\vspace{-1em}\subsubsection{Example MW.}}
\newcommand{\exampleheadAll}{\vspace{-1em}\subsubsection{Examples.}}
\newcommand{\factorhead}{\vspace{-1em}\subsubsection{Factor-graph representation.}}
\newcommand{\altSDP}{dSDP-admm}

%% file: acronyms.tex
\acrodef{ADMM}[ADMM]{alternating direction method of multipliers}
\acrodef{dSDP}[dSDP]{decomposed SDP}
\acrodef{EVR}[EVR]{Eigenvalue ratio}
\acrodef{iSAM}[iSAM]{incremental smoothing and mapping}
\acrodef{GP}[GP]{Gaussian process}
\acrodef{GN}[GN]{Gauss-Newton}
\acrodef{mae}[MAE]{mean absolute error}
\acrodef{NLS}[NLS]{nonlinear least squares}
\acrodef{QCQP}[QCQP]{quadratically-constrained quadratic program}
\acrodef{SDP}[SDP]{semidefinite program}
\acrodef{SLAM}[SLAM]{simultaneous localization and mapping}
\acrodef{STD}[STD]{standard deviation}
\acrodef{CTRO}[CT-RO]{continuous-time range-only localization}
\acrodef{MW}[MW]{matrix-weighted localization}

%% file: _sections/abstract.tex
In recent years, many estimation problems in robotics have been shown to be solvable to global optimality using their semidefinite relaxations.  However, the runtime complexity of off-the-shelf semidefinite programming (SDP) solvers is up to cubic in problem size, which inhibits real-time solutions of problems involving large state dimensions. We show that for a large class of problems, namely those with chordal sparsity, we can reduce the complexity of these solvers to linear in problem size. In particular, we show how to replace the large positive-semidefinite variable with a number of smaller interconnected ones using the well-known chordal decomposition. This formulation also allows for the straightforward application of the alternating direction method of multipliers (ADMM), which can exploit parallelism for increased scalability. We show for two example problems in simulation that the chordal solvers provide a significant speed-up over standard SDP solvers, and that global optimality is crucial in the absence of good initializations.

%% file: _sections/intro.tex
\section{Introduction}\label{sec:intro}

To this date, optimization problems in robotics are predominantly solved with fast local methods. Popular choices in state estimation, for example, are \ac{GN}~\cite{barfoot_state_2017}, or \ac{iSAM}~\cite{kaess_isam_2008,kaess_isam2_2012}; planning and control problems are commonly solved with sequential quadratic programming algorithms and variants thereof~\cite{rawlings_model_2020}.  
Because the cost can become non-convex as soon as non-linear measurement models, dynamic models, or other non-convex constraints are employed, these methods may converge to poor local minima in the absence of well-informed initializations. 

In recent years, important advances have been made towards globally optimal solvers. In computer vision, rotation averaging and registration problems~\cite{kahl_globally_2007,fredriksson_simultaneous_2013,eriksson_rotation_2018} have been globally solved first, followed more recently by methods for optimal outlier-robust estimation~\cite{yang_certifiably_2023}. Classical robotics problems such as range-only localization~\cite{dumbgen_safe_2023,goudar_optimal_2024} and range-aided \ac{SLAM}~\cite{papalia_certifiably_2023}, pose-graph optimization~\cite{rosen_se-sync_2019,carlone_planar_2016}, and isotropic~\cite{holmes_efficient_2023} or non-isotropic~\cite{holmes_semidefinite_2024}~\ac{SLAM} have also seen adaptations of global solvers. We are even seeing a surge of globally optimal solvers for non-smooth planning problems~\cite{marcucci_motion_2023,cohn_noneuclidean_2023}. 

In many of the aforementioned global methods, the computational bottleneck lies in solving a high-dimensional \ac{SDP}. Off-the-shelf solvers for solving \ac{SDP}s commonly use interior-point methods~\cite{nocedal_numerical_2006}, which are of approximately cubic complexity in the number of unknowns and constraints~\cite{andersen_implementing_2003}. This compares favorably with worst-case exponential-time methods such as branch and bound~\cite{olsson_branchandbound_2009}. However, to be competitive with local solvers~\cite{barfoot_state_2017,kaess_isam2_2012}, more efficient \ac{SDP} solvers are desired. Indeed, by exploiting sparsity, state-of-the-art local solvers can run in linear time in either the number of unknown landmarks or poses and, thanks to sophisticated handling of computation graphs, can update solutions efficiently in an online manner~\cite{kaess_isam2_2012}. 

\begin{figure}[tb]
  \centering
  \includegraphics[width=.4\linewidth]{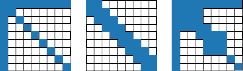}
  \caption{By the nature of variable dependencies, the core matrices of many robotics optimization problems will be chordally sparse. Examples are arrowhead (left) and band-sparse (middle) matrices, or combinations thereof (right). In batch state estimation, local solvers can exploit such sparsity patterns (of the inverse covariance matrix) to yield linear complexity. In this paper, we explain how certifiably optimal SDP solvers can do the same, following problem decomposition and distribution~\cite{zheng_chordal_2021}.} 
  \label{fig:figure1}
\end{figure}

In this paper, we use well-established tools from the optimization community~\cite{boyd_admm_2010,vandenberghe_chordal_2015} to show how \ac{SDP} solvers can also be significantly sped up by exploiting sparsity. For a particular class of problems, namely problems with chordal sparsity (see Figure~\ref{fig:figure1}), the \ac{SDP} can be decomposed into interconnected smaller problems. This reduces the solver complexity and allows the problem to be distributed across computational cores using, for example, the~\ac{ADMM}. We show that such decompositions reduce the solver complexity to approximately linear in the number of variables, which is a significant speedup and comparable to efficient but initialization-dependent local solvers. In summary, our contributions are as follows. \\
\noindent
\text{1.} We adopt the \ac{dSDP} solver and its distributed version from~\cite{zheng_chordal_2021}, and show how they apply to robotics problems, using an intuitive factor-graph language~\cite{dellaert_factor_2021} for modeling. \\
\noindent
\text{2.} We showcase the performance of the methods on two localization problems, which both satisfy the chordal sparsity assumptions, and empirically show that they have the same complexity as efficient \textit{local} solvers. 

We emphasize that the proposed methods have been considered in the control community~\cite{zheng_chordal_2020}, and applied, for example, for optimal power flow~\cite{dallanese_distributed_2013}. Our contribution lies in adopting them for state estimation problems, characterizing their behavior, and, by using the factor-graph language, opening the door to their broader application in robotics. We also pay particular attention to redundant constraints, which are commonly required for tightness in state estimation problems and may significantly impact the effectiveness of the approach. 

This paper is structured as follows. We give an overview of related work and important graph-theoretic identities in Sections~\ref{sec:related_work} and~\ref{sec:background}, respectively. In Section~\ref{sec:method}, we present the methodology to pass from the original sparse non-convex problem to the convex relaxation, design the factor-graph framework to describe and decompose such problems, and finally discuss the sparsity-exploiting solvers. We demonstrate the performance of the two proposed methods on two localization problems in Section~\ref{sec:results} and conclude in Section~\ref{sec:conclusion}.

%% file: _sections/related_work.tex
\section{Related Work}\label{sec:related_work}

The proposed methods lie at the intersection of two active research areas in robotics: efficient certifiably optimal solvers and distributed algorithms.

Methods to speed up the certifiably optimal solvers based on \ac{SDP} relaxations can be broadly categorized into methods that exploit the low-rank property of the sought solution, based on the Burer-Monteiro factorization~\citet{burer_local_2005}, and methods that exploit sparsity of the involved matrices~\citet{vandenberghe_chordal_2015,zheng_chordal_2021}.
The former have been successfully applied to pose-graph optimization, using in particular the Riemannian staircase method~\cite{rosen_se-sync_2019}, and distributed versions of the method have been derived~\cite{tian_distributed_2021,tian_blockcoordinate_2019}. The practicality of the approach has been demonstrated by its adoption in a distributed semantic~\ac{SLAM} framework~\cite{tian_kimeramulti_2022}. However, to this date, the Burer-Monteiro methods have not been successfully applied to problems that require redundant constraints, since they break common constraint qualifications.

Since for many instances in robotics~\cite{yang_certifiably_2023,wise_certifiably_2020,holmes_semidefinite_2024,dumbgen_toward_2024,barfoot_certifiably_2024,goudar_optimal_2024}, redundant constraints are paramount to achieve tight semidefinite relaxations and thus globally optimal solutions, we explore sparsity-exploiting methods as an alternative. On one hand, sparsity can be used to speed up canonical operations within a solver such as solving linear systems of equations: in~\cite{odonoghue_conic_2016}, for example, sparsity of the constraint matrices is used to speed up the projection step of the \ac{ADMM} solver for the homogeneous self-dual embedding problem. Different in nature are so-called decomposition methods for \ac{SDP}s, which are the subject of this paper. Here, the sparsity is exploited to reduce the problem to many smaller-sized instances that can either be solved in one shot~\cite{fukuda_exploiting_2001,kim_exploiting_2011} or in an alternating way \cite{kalbat_fast_2015,sun_decomposition_2014}. We show in this paper that many state estimation problems are suitable candidates for these methods, provided that redundant constraints are such that chordal sparsity is not significantly affected.

The aforementioned decomposition methods lend themselves to distributed implementations. Distribution allows for increased efficiency thanks to parallelism and load sharing, and it has seen a growing interest in robotics in recent years, where real-time requirements and high dimensionality prevail.
In control, for example, there is growing interest in \ac{ADMM}-based quadratic-program solvers~\cite{bambade_proxqp_2022,stellato_osqp_2020}, which lend themselves to distribution~\cite{schubiger_gpu_2020}. In parallel work, a certifiably optimal distributed solver for trajectory optimization, exploiting chain-like sparsity as a special case of correlative sparsity~\cite{lasserre_convergent_2006}, was developed~\cite{kang_fast_2024}. In estimation, for high-dimensional problems such as structure from motion and bundle adjustment, distributed methods based on majorization minimization techniques~\cite{fan_decentralization_2022} or Gaussian belief propagation~\cite{ortiz_bundle_2020} have shown impressive scalability improvements. Distributed algorithms can also be shared across multiple robots, allowing for collaborative solutions to, for example, localization and mapping~\cite{carlone_simultaneous_2011,choudhary_distributed_2017,mcgann_asynchronous_2024}, with dedicated measures to address limited communication bandwidth and proximity constraints~\cite{cunningham_ddfsam_2010,cieslewski_data-efficient_2018,murai_distributed_2024}. Since they distribute a \emph{local} solver, and often solve non-convex problems, the above methods may converge to poor local minima. The \ac{ADMM} algorithm described herein can be seen as a first step towards a distributed \emph{and} certifiably optimal alternative for robotics.

%% file: _sections/background.tex
\section{Background}\label{sec:background}

\subsection{Notation and definitions}

We denote scalars, vectors, matrices, and sets, respectively, by the fonts $x$, \vc{x}, \vc{X}, and $\mathcal{X}$. The $j$-th element of a vector, the $j$-th column of a matrix, and the element $(i,j)$ of a matrix are denoted, respectively, by $\vc{x}[j]$, $\vc{X}[j]$, and $\vc{X}[i,j]$. We write $\vc{X}\succeq 0$ to mean \vc{X} is positive semidefinite. $\vc{I}_d$ and $\vc{0}_d$ are the identity matrix and vector of all zeros of size $d$, respectively, and $d$ is dropped when clear from context. The operator $\vech(\cdot)$ returns the upper-diagonal elements of a symmetric matrix $\vc{X}$, where off-diagonal elements are multiplied by $\sqrt{2}$, ensuring $\tr{\vc{Q}^\top\vc{X}}=\vechh{(\vc{Q})}^\top\vech{(\vc{X})}$. The inverse operation is $\mat(\cdot)$: $\vc{X} = \mat{(\vc{y})}$ and we write $\vc{y}\succeq 0$ for $\mat{(\vc{y})} \succeq 0$. An undirected graph is defined as $\mathcal{G}=(\mathcal{V},\mathcal{E})$, where $\mathcal{V}$ is the set of vertices and $\mathcal{E}$ is the set of edges. We use $(i,j)\in\mathcal{E}$ and $(i,j)\in\mathcal{G}$ interchangeably, as well as $i\in\mathcal{V}$ and $i\in\mathcal{G}$. A \textit{subgraph} induced by a set of vertices $\mathcal{W}\subset\mathcal{V}$ is the graph with vertices $\mathcal{W}$ and edges $\mathcal{E}\cap(\mathcal{W}\times\mathcal{W})$. A \textit{clique} is a subgraph that is fully connected. A \textit{cycle} is a set of pairwise distinct vertices ${v_1, \ldots, v_k}\subset\mathcal{V}$ such that $(v_1, v_k) \in \mathcal{E}$ and $(v_i, v_{i+1})\in\mathcal{E}$. A \textit{chord} in a cycle is an edge connecting two non-consecutive vertices.

\subsection{Chordal sparsity in SDPs}\label{sec:chordal}

An extensive treatment of sparsity in \ac{SDP}s is given in~\cite{zheng_chordal_2021}. Here, we introduce some lighter notation and highlight the concepts that will be the most useful to us in the remainder of the paper. Given a set of symmetric matrices $\mathcal{A}=\{\vc{A}_1, \ldots, \vc{A}_K\}$, we define the \textit{aggregate sparsity pattern} of $\mathcal{A}$ as the following set of pairwise indices:
\begin{equation}
  \mathcal{E} = \{(i, j)\, | \,\exists \,\bm{A}_k\in\mathcal{A} \text{ s.t. } \bm{A}_k[i,j]\neq 0\}.
\end{equation}
In other words, this is the set of all index pairs at which at least one of the matrices in the set has a non-zero element. We define the \textit{sparsity graph} of $\mathcal{A}$ as $\mathcal{G}=(\{1,\ldots,N\}, \mathcal{E})$. Next, we reiterate a few well-known properties of graphs.

\vspace{-0.2cm}
\begin{definition}[Chordal graph]
  A graph $\mathcal{G}$ is chordal if every cycle of length $\geq 4$ contains at least one chord. 
\end{definition}

A graph can be rendered chordal by adding edges until it is chordal. Finding the minimum chordal extension, i.e., the minimum set of chords to render the graph chordal, is an NP-hard problem, but efficient heuristics exist~\cite{amestoy_approximate_1996}. In this paper, we will use specific sparsity patterns that typically arise in robotics applications and use deterministic chordal extensions.

\vspace{-0.2cm}
\begin{definition}[Maximal clique]
  $\mathcal{C}$ is a maximal clique of $\mathcal{G}$ if it is not contained in any other clique of $\mathcal{G}$.
\end{definition}
\vspace{-0.2cm}

Finally, we say that a matrix is semidefinite-completable in $\mathcal{E}$, written as $\vc{X}_{\mathcal{E}}\succeq 0$, if \vc{X} can be made positive semidefinite by changing only the elements corresponding to indices \textit{not} in $\mathcal{E}$.

Now, we can state the important property that enables us to replace the large \ac{SDP} variable with smaller ones. We consider~\ac{SDP}s of the general form
  \begin{equation}
  \min\{\tr{\vc{Q}\vc{X}} | \vc{X}\succeq0,\tr{\vc{A}_i\vc{X}}=b_i,i=1\ldots K\},
\end{equation}
where $\bm{Q}$ is a symmetric cost matrix. Then we have the following property:

\vspace{-0.2cm}
\begin{thm}{\cite[Theorem 2.2]{zheng_chordal_2021}}\label{thm}
  Let $\mathcal{E}_\text{SDP}$ be the aggregate sparsity pattern of the set of problem matrices, $\{\vc{Q},\vc{A}_1,\ldots,\vc{A}_K\}$, and let $\mathcal{G}_{\text{SDP}}$ be the sparsity graph corresponding to its chordal extension. Let the maximal cliques of $\mathcal{G}_{\text{SDP}}$ be 
  $\mathcal{C}_1, \ldots, \mathcal{C}_T$ and let $\vc{C}_1, \ldots, \vc{C}_T$ be the submatrices of \vc{X} at the indices of the cliques. Then we have: $\vc{X}_{\mathcal{E}}\succeq 0\iff (\forall t)\, \vc{C}_t\succeq 0$.
\end{thm}
\vspace{-0.2cm}

%% file: _sections/method.tex
\section{Method}\label{sec:method}

In this section, we present two scalable methods for solving localization problems in robotics to global optimality, based on~\cite{zheng_chordal_2021}, which rely on the graph-theoretic concepts outlined in the previous section. We formally define the state estimation problem and its reformulation to a~\ac{QCQP} in Sections~\ref{sec:nls} and~\ref{sec:qcqp}, respectively. We then derive the monolithic~\ac{SDP} in~\ref{sec:sdp} and show how it can be decomposed and distributed in Sections~\ref{sec:dsdp} and~\ref{sec:dsdp-admm}. A summarizing diagram of the different reformulations is given below.

\vspace{0em}
\begin{tikzpicture}[every node/.style={font=\scriptsize}]
  \node [anchor=north,minimum height=0.5cm] (note1) at (0,0) {NLS};
  \node [anchor=south, align=center] (exp1) at (1,-0.1) {\strut substitutions};
    \node [anchor=north, align=center] (exp1) at (1,-0.3) {{4.2}};
    \node [anchor=north,minimum height=0.5cm] (note2) at (2.2,0) {QCQP};
    \node [anchor=south, align=center] (exp2) at (3.3,-0.1) {rank \\ \strut relaxation };
    \node [anchor=north, align=center] (exp2) at (3.4,-0.3) {{4.3}};
    \node [anchor=north,minimum height=0.5cm] (note3) at (4.4,0) {SDP};
    \node [anchor=south, align=center] (exp3) at (5.4,-0.1) { chordal \\ \strut decomposition};
    \node [anchor=north, align=center] (exp3) at (5.4,-0.3) {{4.4}};
    \node [anchor=north,minimum height=0.5cm] (note4) at (6.6,0) {dSDP};
    \node [anchor=south, align=center] (exp4) at (7.9,-0.1) {operator \\ \strut splitting};
    \node [anchor=north, align=center] (exp4) at (7.9,-0.3) {{4.5}};
    \node [anchor=north,minimum height=0.5cm] (note5) at (9.6,0) {dSDP-admm};
    \draw[->] (note1.east) -- (note2.west);
    \draw[->] (note2.east) -- (note3.west);
    \draw[->] (note3.east) -- (note4.west);
    \draw[->] (note4.east) -- (note5.west);
\end{tikzpicture}
\vspace{-1em}

\subsection{Nonlinear least squares (NLS) formulation}\label{sec:nls}
Many classic state estimation problems in robotics, including localization and~\ac{SLAM}, take the form of a~\ac{NLS} problem when we seek the maximum a-posteriori estimator~\cite{barfoot_state_2017}:
\begin{equation}
  \begin{aligned}
    \hat{\vc{\xi}} = \argmin{\vc{\xi} \in \mathcal{X}} \sum_{k \in \mathcal{Q}} &\underbrace{\vc{e}_{k}(\vc{\xi}_k)^\top\vc{W}_{k}\vc{e}_{k}(\vc{\xi}_k)}{a_k(\vc{\xi}_k)} 
    + \sum_{(i,j)\in\mathcal{R}} \underbrace{\vc{e}_{ij}(\vc{\xi}_i, \vc{\xi}_j)^\top \vc{W}_{ij}\vc{e}_{ij}(\vc{\xi}_i, \vc{\xi}_j)}{r_{ij}(\vc{\xi}_i, \vc{\xi}_j)},
  \end{aligned}
  \label{eq:nls}
\end{equation}
where $\vc{\xi}=\{\vc{\xi}_1, \ldots, \vc{\xi}_N\}$ represents the state variables to be estimated. We denote by $\mathcal{Q}$ ($\mathcal{R}$) the set of indices (pairs) at which we have absolute (relative) measurements or prior terms. The absolute and relative cost terms are denoted $a_k(\vc{\xi}_k)$ and $r_{ij}(\vc{\xi}_i, \vc{\xi}_j)$, respectively, and defined by their respective error terms $\vc{e}_{.}(\cdot)$ and corresponding inverse covariance matrix $\vc{W}_{.}$.

Note that, for non-linear $\vc{e}_k$, $\vc{e}_{ij}$, or non-convex $\mathcal{X}$, Problem~\eqref{eq:nls} may be non-convex and thus its globally optimal solutions may be hard to find efficiently. Practitioners commonly resort to local solvers such as~\ac{GN}~\cite{barfoot_state_2017} or~\ac{iSAM}~\cite{kaess_isam_2008}, which are efficient but have the important caveat that they may converge to poor local minima in the absence of a good initialization.

\factorhead Problem~\eqref{eq:nls} can be represented by a factor graph $\mathcal{G}_f=(\mathcal{V},\mathcal{M},\mathcal{E})$~\cite{dellaert_factor_2021}, where $\mathcal{V}$ contains all states, $\mathcal{M}$ contains the measurements or priors, and $\mathcal{E}$ defines the (bipartite) connectivity between $\mathcal{V}$ and $\mathcal{M}$. In this paper, we are interested in factor graphs that are chordally sparse -- in other words, graphs that have small maximum cliques. All examples given in Figure~\ref{fig:figure1} fall into this category. Throughout this paper, we will consider two localization problems as examples where $\mathcal{E}$ is a chain. Their sparsity patterns are given in Figure~\ref{fig:factor-graphs}.\footnote{Note that we are not limited to chains, but chains are a particularly well-behaved structure with uniformly sized, small maximum cliques.} We define these problems next, represented by the respective error terms entering~\eqref{eq:nls}.

\exampleheadRO We adopt the \exampleRO{} problem from~\cite{dumbgen_safe_2023}, where our goal is to find the position of a moving device over time, given distance measurements to $N_m$ fixed and known landmarks. 
Because distance measurements may be sparse, we add a~\ac{GP} motion prior to encourage smoothness and to ensure that recovery is possible even in underdetermined cases~\cite{barfoot_batch_2014}. 
It was found in~\cite{dumbgen_safe_2023} that local solvers are prone to converge to local minima for this application, particularly when the landmarks are in degenerate configurations. The state is given by $\vc{\xi}_i = \vc{x}_i = \bmat{\vc{t}_{i}^\top & \vc{v}_{i}^\top}^\top$, where $\vc{t}_i$ and $\vc{v}_i$ are the position and velocity at time index $i$, respectively.\footnote{For conciseness, we focus on the constant-velocity motion prior in this paper, but the extension to other priors, such as constant-acceleration or zero-velocity, is straightforward~\cite{dumbgen_safe_2023}.} The motion-prior errors are given by $\vc{e}_{ij} = \vc{x}_i - \vc{\Phi}_{ij}\vc{x}_j$, where $\vc{\Phi}_{ij}$ is the so-called transition matrix, and $\vc{W}_{ij}$ is calculated using the \ac{GP} motion-prior covariance~\cite{barfoot_batch_2014}. Measurement error terms are given by $\vc{e}_{k}\inR{N_k}$ with $N_k$ the number of landmarks seen at time index $k$, and the $i$-th element given by $\vc{e}_{k}[i] = \tilde{d}_{ki}^2 - \norm{\vc{t}_k - \vc{m}_i}^2$,
where $\vc{m}_i$ is the $i$-th known landmark position, and $\tilde{d}_{ki}$ is the corresponding distance measurement. The matrix $\vc{W}_k$ is set to the assumed inverse covariance matrix of the squared distance measurements.

\exampleheadMW In \exampleMW, adopted from~\cite{holmes_semidefinite_2024}, we seek to estimate the pose of a moving device based on Euclidean observations of known landmarks. Because such observations typically come from sensors with varying accuracy in the radial and lateral directions, it is often crucial to allow for matrix-weighted, \ie, anisotropic, measurement noise. It was found in~\cite{holmes_semidefinite_2024} that this problem exhibits local minima, particularly as measurement anisotropicity increases.
We define the state by $\vc{\xi}_i: \vc{T}_{i0}\in\SE{3}$, composed of 
rotation $\vc{C}_{i0}\in\SO{3}$, and translation $\vc{t}_{0i}^{i} \inR{3}$, as explained in more detail in~\cite{holmes_semidefinite_2024}.
Given this state definition, the relative error terms can be expressed as $\vc{e}_{ij} = \text{vec}(\vc{T}_{i0} - \widetilde{\vc{T}}_{ij} \vc{T}_{j0})$, where $\widetilde{\vc{T}}_{ij}$ is a measurement of the relative transform from pose $i$ to pose $j$. For relative measurements, we assume isotropic noise, thus $\vc{W}_{ij}=\sigma_{ij}^{-2} \vc{I}$. For absolute state measurements, we define
$\vc{e}_{k,l} = \widetilde{\vc{m}}_{lk}^k - \vc{T}_{k0} \vc{m}_{l0}^0$, where $\widetilde{\vc{m}}_{lk}^k$ is the Euclidean measurement of landmark $l$ as seen from pose $k$, and $\vc{m}_{l0}^0$ are the known landmark positions, expressed in world frame, and $\bm{W}_k$ is a typically anisotropic inverse covariance matrix.

\subsection{NLS to Quadratically Constrained Quadratic Program}\label{sec:qcqp}
In many applications of interest, Problem~\eqref{eq:nls} can be written as a polynomial optimization problem, which in turn can be formulated as an equivalent~\ac{QCQP}. The intuition is that we introduce a `lifted' vector \vc{x} that contains \vc{\xi} and potentially some higher-order `lifting functions' thereof -- enough so that both the cost and the constraints stemming from $\mathcal{X}$ become quadratic in \vc{x}. 

More specifically, we call the lifting functions $\vc{\ell}_k(\vc{\xi}_k)$ and introduce $h=1$, the so-called `homogenization' variable, which ensures that linear and constant cost terms can also be written in quadratic form. Introducing the vectors\footnote{We include one homogenization variable per variable group for simplicity of exposition. In practice, the homogenization variable can be shared among different groups.} 
\begin{equation}
  \vc{x}_k := \bmat{h & \vc{\xi}_k^\top & \vc{\ell}_k(\vc{\xi}_k)^\top}^\top, \quad
  \vc{x}_{ij} := \bmat{h & \vc{\xi}_i^\top & \vc{\ell}_i(\vc{\xi}_i)^\top
  & \vc{\xi}_j^\top & \vc{\ell}_j(\vc{\xi}_j)^\top}^\top,
  \label{eq:xik}
\end{equation}
the~\ac{QCQP} reformulation of~\eqref{eq:nls} takes the form:
\begin{equation}
  \begin{aligned}
    \hat{\vc{x}} = \argmin{\vc{x}} &\sum_{k\in\mathcal{Q}} \vc{x}_k^\top\vc{Q}_k\vc{x}_k + \sum_{(i, j)\in\mathcal{R}}\vc{x}_{ij}^\top\vc{R}_{ij}\vc{x}_{ij} \\
    \text{s.t. }
    &\vc{x}_k^\top \vc{A}_{j_k}^s \vc{x}_k = b_{j_k}, \quad k\in\mathcal{V},j_k=1\ldots N_{k}^s \\
    &\vc{x}_k^\top \vc{A}_{j_k}^p \vc{x}_k = 0, \quad k\in\mathcal{V},\,j_k=1\ldots N_{k}^p 
  \end{aligned},
  \label{eq:qcqp-split}
\end{equation}
where $N_k^p$ is the number of primary constraints of node $k$ (from $\mathcal{X}$) and $N_k^s$ is the number of substitution constraints. The link between the cost matrices $\vc{Q}_k$ and $\vc{R}_{ij}$ and the original error terms in~\eqref{eq:nls}, as well as the form of the substitution and primary constraint matrices $\bm{A}_{jk}^s, \bm{A}_{ik}^p$, is given in Appendix~\ref{app:general}.\footnote{Note that each primary constraint is only a function of one variable group, akin to the separability assumption assumed in related works~\cite{zheng_chordal_2021}, which is crucial for efficient problem distribution.}

\begin{figure}[tb]
\centering
\begin{minipage}{.49\linewidth}
  \centering
  \footnotesize{NLS factor graph $\mathcal{G}_f$} \\
  \includegraphics[width=.8\linewidth]{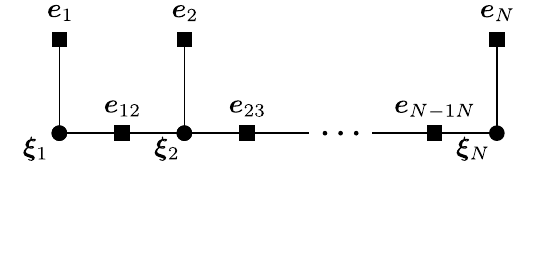}
\end{minipage}
\begin{minipage}{.49\linewidth}
  \centering
  \footnotesize{QCQP factor graph  $\mathcal{G}_q$} \\
  \includegraphics[width=.8\linewidth]{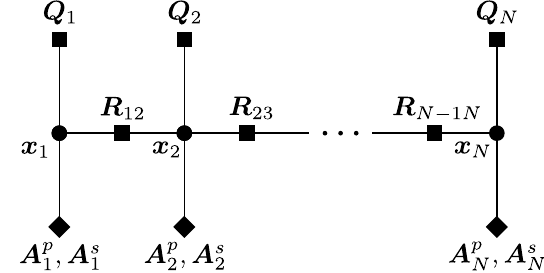}
\end{minipage}
\caption{Factor graphs~\cite{dellaert_factor_2021} for the localization problems treated in this paper. 
On the left is the factor graph of the original \ac{NLS} problem~\eqref{eq:nls}. On the right is the variant of factor graphs proposed to describe the~\ac{QCQP} reformulation~\eqref{eq:qcqp-split}. This graph is formulated regarding the lifted states $\vc{x}_k$ and includes constraint and cost factors. Round nodes are variables, square nodes are cost terms, and diamonds are constraints.}\label{fig:factor-graphs}
\end{figure}

\factorhead We represent the reformulated Problem~\eqref{eq:qcqp-split} as a variant of factor graphs that includes constraints on variables, which has also been considered in~\cite{cunningham_ddfsam_2010,qadri_incopt_2022}. We call this factor graph $\mathcal{G}_q$, shown in Figure~\ref{fig:factor-graphs}, where constraint factors are represented with diamonds. The factor graph can be seen as the generalization of $\mathcal{G}_f$  (which represents the cost of~\eqref{eq:nls}) to the Lagrangian of~\eqref{eq:qcqp-split}. While there is a one-to-one mapping from the connectedness of the factor graph $\mathcal{G}_f$ to the sparsity of the information matrix of the original problem~\eqref{eq:nls}~\cite{dellaert_factor_2021}, the connectedness of the factor graph $\mathcal{G}_q$ reflects the aggregate sparsity of the matrices $\vc{Q}$, $\vc{A}_{j_k}^{s}$ , and $\vc{A}_{j_k}^{p}$ in~\eqref{eq:qcqp-split}. 

\exampleheadAll For brevity, we provide the details of the introduced lifting functions for the~\ac{CTRO}~and~\ac{MW} examples in Appendix~\ref{app:examples}.  The sparsity patterns of the problem matrices for the two examples are provided in Figure~\ref{fig:agg_sparsity}. 

\subsection{QCQP to Semidefinite Program}\label{sec:sdp}

We first derive the monolithic semidefinite relaxation of~\eqref{eq:qcqp-split}, which is our centralized baseline. To this end, we introduce the combined vector $\vc{x}\inR{N_x}$,
\begin{equation}
\vc{x} := \bmat{h & \vc{\xi}_1^\top & \vc{\ell}_1(\vc{\xi}_1)^\top 
& \cdots & \vc{\xi}_N^\top & \vc{\ell}_N(\vc{\xi}_N)^\top}^\top,
\end{equation}
and the projection matrices $\vc{E}^k$, $\vc{E}^{ij}$, defined implicitly through $\vc{x}_k=\vc{E}^k\vc{x}$, and $\vc{x}_{ij}=\vc{E}^{ij}\vc{x}$. Then, we rewrite~\eqref{eq:qcqp-split} as
\begin{equation}
  \begin{aligned}
    \hat{\vc{x}} = \argmin{\vc{x}}  
		 &\vc{x}^\top{\vc{Q}} \vc{x} + \vc{x}^\top{\vc{R}}\vc{x} \\
    \text{s.t. } 
    &\vc{x}^\top \widebar{\vc{A}}_{j_k}^s \vc{x} = b_{j_k}, \quad k\in\mathcal{V},j_k=1\ldots N_k^s \\
    &\vc{x}^\top \widebar{\vc{A}}_{j_k}^p \vc{x} = 0, \quad k\in\mathcal{V},j_k=1\ldots N_k^p
  \end{aligned},
  \label{eq:qcqp-full}
\end{equation}
with $\vc{Q}=\sum_k\widebar{\vc{Q}}_k$ and $\vc{R}=\sum_{ij}\widebar{\vc{R}}_{ij}$. The constraint matrices, $\widebar{\vc{A}}_{j_k}^p,\widebar{\vc{A}}_{j_k}^s$, and the cost matrix $\widebar{\vc{Q}}_k$ in~\eqref{eq:qcqp-full}, are obtained using $\widebar{\vc{Y}}_k=\vc{E}^k\vc{Y}_k\vc{E}^{k\top}$. Similarly, we have defined $\widebar{\vc{R}}_{ij}=\vc{E}^{ij}\vc{R}_{ij}\vc{E}^{ij\top}$.

Every QCQP can be relaxed to an \ac{SDP} by using the succession of identities $\bm{x}^\top\bm{Q}\bm{x} = \tr{\bm{x}^\top\bm{Q}\bm{x}}= \tr{\bm{Q}\bm{x}\bm{x}^\top} = \{ \tr{\bm{Q}\vc{X}} | \vc{X}=\vc{x}\vc{x}^\top \} =\{ \tr{\bm{Q}\vc{X}} | \vc{X}\succeq 0, \rank{\vc{X}}=1\}$. By relaxing the rank-one constraint, we obtain the following primal relaxation of~\eqref{eq:qcqp-full}:
\begin{equation}
  \begin{aligned}
    \hat{\vc{X}} = \argmin{\vc{X}\succeq 0} &\tr{\vc{Q}\vc{X}}+\tr{\vc{R}\vc{X}} \\
    \text{s.t. } 
    &\tr{\bar{\vc{A}}_{j_k}^s \vc{X}} = b_{j_k}, \quad k\in{\mathcal{V}}, j_k=1\ldots N_k^s \\
    &\tr{\bar{\vc{A}}_{j_k}^p \vc{X}} = 0, \quad k\in\mathcal{V}, j_k=1\ldots N_k^p \\
    &\tr{\bar{\vc{A}}_{i}^r \vc{X}} = 0, \quad i=1\ldots N_r
  \end{aligned},
  \label{eq:sdp}
\end{equation}
\noindent
where we have introduced $\bar{\bm{A}}^r_{i}$, so-called redundant constraints. Redundant constraints are trivially satisfied constraints of~\eqref{eq:qcqp-full}, \ie, they are linearly dependent in~\eqref{eq:qcqp-full} but are independent in~\eqref{eq:sdp}. Their role is to help enforce the structure that may be lost when dropping the rank constraint, thus encouraging the solution $\hat{\vc{X}}$ to be of low rank. This is important because only if the optimal solution $\hat{\vc{X}}$ of~\eqref{eq:sdp} is of rank one can we obtain the globally optimal solution $\hat{\vc{x}}$ of~\eqref{eq:qcqp-full} from the factorization $\hat{\vc{X}}=\hat{\vc{x}}\hat{\vc{x}}^\top$ (in which case we say that the relaxation is \textit{tight}).

\begin{figure}[tb]
  \centering
  \includegraphics[width=\linewidth]{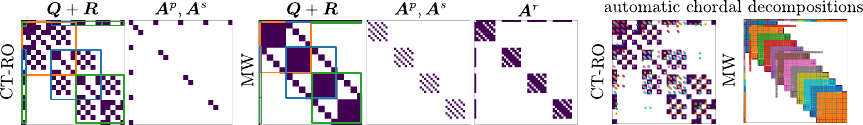}
  \caption{Sparsity patterns for \ac{CTRO} and \ac{MW}. We show the aggregate sparsity of the cost matrix $\vc{Q}+\vc{R}$, and the primary, substitution, and redundant constraints: $\vc{A}^p$, $\vc{A}^s$, and $\vc{A}^r$. The overlaid colored boxes on the cost matrix mark our used chordal decomposition of the aggregate sparsity patterns. The right-most plots show automatically generated chordal decompositions (different colors mark different cliques).}\label{fig:agg_sparsity}
\end{figure}

We end this section by making a few observations about Problem~\eqref{eq:sdp}. 
The constraint $\bm{X}\succeq 0$ links variables that otherwise would not interact through the cost or constraint matrices. However, Theorem~\ref{thm} suggests that if the measurement graph $\mathcal{G}_p$ is chordally sparse, we can regain efficiency by splitting the problem into its maximal cliques. 
Note that the redundant constraints $\bm{A}^r_{i}$, which may be required for tightness, can however induce new dependencies in $\mathcal{G}_p$ and reduce sparsity (\ie, increase the size of the maximal cliques). Thus, we deal with opposing forces: without redundant constraints, the \ac{SDP} relaxation may not be tight, preventing us from getting a globally optimal estimate. However, adding redundant constraints may reduce sparsity and prevent us from effectively solving the \ac{SDP}. As we see next, the two example problems have tight semidefinite relaxations with essentially unchanged sparsity patterns.

\exampleheadRO For this problem, the relaxation is already tight without the need for redundant constraints, as observed in~\cite{dumbgen_safe_2023} and~\cite{dumbgen_toward_2024}. 

\exampleheadMW For this problem, we can add the following well-known redundant constraints~\cite{briales_convex_2017}: $\vech{(\vc{C}_{i0})}^\top\vech{(\vc{C}_{i0}^\top)}=\vech{(\vc{I}_3)}$\footnote{We use only 5 of these 6 constraints since one is linearly dependent.} and the redundant handedness constraints $\vc{C}_{i0}[2]\times\vc{C}_{i0}[3]=\vc{C}_{i0}[1]$ and $\vc{C}_{i0}[3]\times\vc{C}_{i0}[1]=\vc{C}_{i0}[2]$. It was shown in~\cite{holmes_semidefinite_2024} that these redundant constraints may significantly improve tightness as the noise level increases and the number of measurements decreases. Since the constraints depend only on one pose at a time, they do not change the structure of the measurement graph, which can also be observed from the sparsity patterns in Figure~\ref{fig:agg_sparsity}.

%% file: _sections/algos.tex
\subsection{Decomposed solution: dSDP}\label{sec:dsdp}

We can now describe our monolithic solver, which decomposes the problem into smaller parts but runs in a centralized way. We adopt a vectorized~\ac{SDP} formulation from here on for notational convenience. In particular, we introduce variable $\vc{y} = \vech{(\vc{X})}$ and $\vc{y}_k=\vech{(\vc{X}_k)}$. Using this notation, we can write~\eqref{eq:sdp} in the standard dual form for conic programs~\cite{boyd_convex_2004}:
\begin{equation}
  \begin{aligned}
    \hat{\vc{y}} = \argmin{\vc{y}\succeq 0} &\sum_{k\in\mathcal{Q}} \tilde{\vc{q}}_k^\top\vc{y}_k + \sum_{i,j\in\mathcal{R}} \tilde{\vc{r}}_{ij}^\top\vc{y}_{ij} \\
    \text{s.t. } & \vc{A}_n \vc{y}_n = \vc{b}_n, \quad n \in \mathcal{V}
  \label{eq:sdp-vec}
\end{aligned},
\end{equation}
where we have introduced $\tilde{\vc{q}}_k:=\vechh{(\vc{Q}_k)}$, $\tilde{\vc{r}}_{ij}:=\vechh{(\vc{R}_{ij})}$, and we have grouped all primary, substitution, and redundant constraints into $\vc{A}_n$: \begin{equation}
  \begin{aligned}
  \vc{A}_n^\top &:= \bmat{
  \cdots
  \vechh{(\vc{A}^p_{j_k})}
  \cdots
  \vechh{(\vc{A}^s_{j_k})}
  \cdots
  \vechh{(\vc{A}^r_{j_k})}
  \cdots
  }, 
  \end{aligned}
  \label{eq:Ab}
\end{equation}
and $\vc{b}_n$ is defined accordingly. Typically, the lack of speed in~\ac{SDP}s is due to the semidefinite constraint on $\vc{y}$. However, we can now apply Theorem~\ref{thm}, allowing us to decompose this constraint into a sequence of smaller positive-definiteness constraints. Indeed, if the aggregate sparsity pattern of the~\ac{SDP} is chordally sparse, we can replace the condition $\vc{y}\succeq 0$ with $\vc{c}_t\succeq0, t=1\ldots T$ where $\vc{c}_t$ correspond to the variables from the clique $t$.

Given the aggregate sparsity pattern of Problem~\eqref{eq:sdp}, we could automatically compute the maximal cliques using efficient heuristics~\cite{amestoy_approximate_1996}. Figure~\ref{fig:agg_sparsity}, on the right, gives example decompositions using the approximate minimum degree ordering. We observe that the method has generated some fill-in between variables groups that do not have dependencies in the original graph $\mathcal{G}_q$. This reduces the sparsity, which would affect our solver's computational cost. Instead, we suggest using a manual decomposition, which, although potentially less minimal, avoids such dependencies and generates fewer cliques. Indeed, a straightforward decomposition of the problem is given by $\bm{c}_i=(h, \bm{y}_{i},\bm{y}_{i+1})$, also shown in Figure~\ref{fig:agg_sparsity} for the same example. For simplicity, we use the trivial maximal chordal extension within each clique. Exploiting chordal sparsity within each clique is expected to only lead to a minor improvement.\footnote{This manual decomposition also works for graphs that are not simple chains: we can first find a chordal decomposition at the variable block level and then use the maximal chordal extension within each block.}

More formally, we define the {\textit{chordal clique tree}} $\vc{T}_c:={(\mathcal{V}_c,\mathcal{E}_c)}$, whose vertices index the cliques, 
and edges between two cliques indicate that they share variables. This is necessarily a tree because any loop closure in this graph would violate the definition of maximal cliques. Using this notion and Theorem~\ref{thm}, we can define the (dSDP) problem:
\begin{equation}
  \begin{aligned}
    \{\hat{\vc{c}}_1, \ldots, \hat{\vc{c}}_T\} = \argmin{\vc{c}_1, \ldots, \vc{c}_T} & \sum_{t\in\mathcal{V}_c} \vc{q}_t^\top\vc{c}_t \\
    \text{s.t. } & \vc{A}_t \vc{c}_t = \vc{b}_t, \quad t\in\mathcal{V}_c \\
    & \vc{S}_t^{t'} \vc{c}_t = \vc{S}_{t'}^t\vc{c}_{t'},  \quad t,t' \in \mathcal{E}_c  \\
    & \vc{c}_t\succeq 0, \quad t\in\mathcal{V}_c
  \label{eq:dsdp}
  \end{aligned},
\end{equation}
where $\vc{S}_t^{t'}$ is the indexing matrix selecting the shared variables with $t'$ from clique $t$, and $\vc{q}_t$ is the clique-wise cost vector, obtained from splitting the cost variables equally across overlapping nodes. 
Problem~\eqref{eq:dsdp} can be solved by any off-the-shelf~\ac{SDP} solver.

\begin{figure}[tb]
  \centering
  \includegraphics[width=.8\linewidth,trim={0 1cm 0 0},clip]{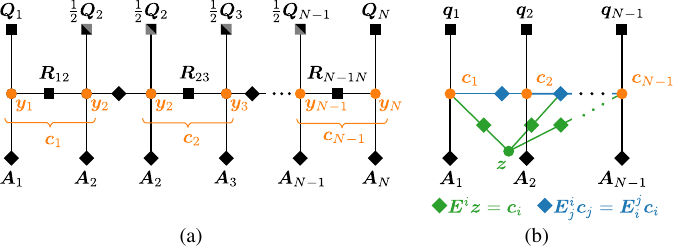}
  \caption{Visualization of the decomposition of the chordal localization problems. On the left is the factor graph split into maximal cliques, with the shared cost terms $\vc{Q}_i,i=1\ldots N-1$ split into half, and constraints added to enforce the overlap of neighboring cliques. On the right is the corresponding clique tree, used for \ac{dSDP} (constraints in blue) and \altSDP~(consensus variable $\vc{z}$ and constraints shown in green).}
  \label{fig:dfg}
\end{figure}

\factorhead The process of splitting variables and associated cost terms is visualized, for our example problems, in the factor graph in Figure~\ref{fig:dfg}(a). We also provide a sketch of the associated clique tree, which includes the introduced overlap constraints in~\eqref{eq:dsdp}. Note that the proposed decomposition has the advantage of having a clean factor-graph representation since it keeps the original variable groups intact.

\vspace{1em}
Problem~\eqref{eq:dsdp} has the potential of being significantly faster to solve because the large \ac{SDP} variable is split into $T:=|\mathcal{V}_c|$ smaller ones. As the complexity of interior-point solvers depends on both the problem dimension and the number of constraints~\cite{andersen_implementing_2003}, the speedup is a function of the number of cliques, their size, and their connectivity. We will see that this speed-up is significant for the studied localization problems and our manual decomposition. Note that Problem~\eqref{eq:dsdp} is still solved in a centralized manner as the overlapping constraints introduce dependencies between the cliques. Next, we formulate the distributed algorithm.
 
\subsection{Distributed solution: dSDP-admm}\label{sec:dsdp-admm}

In this section, we state a distributed version of~\ac{dSDP} using~\ac{ADMM}, which we call~\altSDP. 
Our method is an adaptation of the so-called consensus \ac{ADMM}~\cite[Chapter 7]{boyd_admm_2010}, 
where one variable group contains local but overlapping variables, and the other variable group contains the `consensus' among overlapping variables. We create the feasible set $\mathcal{F}_t := \{\vc{c}_t | \vc{c}_t\succeq 0, \vc{A}_t\vc{c}_t = \vc{b}_t\}$ and define the indicator function $\mathcal{I}_\mathcal{X}(\vc{x})$ which maps to 0 when $\vc{x}\in\mathcal{X}$ and to $\infty$ otherwise. Then, we can reformulate~\eqref{eq:dsdp} as follows:

\begin{equation} \begin{aligned}
    \min_{\vc{c}_1, \ldots, \vc{c}_T} & \sum_{t\in\mathcal{V}_c} \vc{q}_t^\top\vc{c}_t + \mathcal{I}_{\mathcal{F}_t}(\vc{c}_t) \\
    \text{s.t. } &\vc{S}^{t} \vc{z} = \vc{c}_{t}, \quad t\in\mathcal{V}_c
  \label{eq:new}
\end{aligned},
\end{equation}
\noindent where we have introduced the consensus variable \vc{z}, similar to \vc{x} in the original problem. Importantly, the semi-definite constraint is now on the smaller variables $\vc{c}_t$. The clique tree corresponding to this problem is shown in Figure~\ref{fig:dfg}(b).

Applying the standard \ac{ADMM} iterations to~\eqref{eq:new}, we get: 
\newcommand{\errterm}{\vc{S}^t\vc{z}^i - \vc{c}_t}
\begin{subequations}
  \begin{align}
    (\forall t):\, \vc{c}_t^{i} &= \argmin{\vc{c}_t} \mathcal{L}_\rho(\vc{c}_t, \vc{z}^i, \vc{\lambda}_t^i),\label{eq:admm-step1} \\
    \vc{z}^{i} &= \argmin{\vc{z}} \mathcal{L}_\rho(\vc{c}^{i}, \vc{z}, \vc{\lambda}^i),\label{eq:admm-step2} \\
    (\forall t):\, {\vc{\lambda}_t}^{i+1} &= {\vc{\lambda}}_t^i + \rho\left(\errterm^i\right),\label{eq:admm-step3}
  \end{align}
\end{subequations}
\newcommand{\errtermh}{\vc{S}^t\vc{z} - \vc{c}_t}
where the augmented Lagrangian is given by $\mathcal{L}_{\rho}(\vc{c}, \vc{z}, \vc{\lambda}) = f(\vc{c}) + g(\vc{z}) + \sum_t\vc{\lambda}_t^\top(\errtermh) + \frac{\rho}{2}\norm{\errtermh}^2$. Thanks to the introduction of $\vc{z}$, step~\eqref{eq:admm-step1} is decomposed into $T$ optimization problems, each of which is an~\ac{SDP} with quadratic cost that can reformulated to a linear~\ac{SDP} using a linear matrix inequality, see \eg,~\cite{madani_admm_2015}.  
The second step~\eqref{eq:admm-step2} is an unconstrained problem with a closed-form solution: each element of $\vc{z}$, consists of the average the cliques which depend on it. Therefore, $\vc{z}$ can also be updated in a distributed manner, where current values of $\vc{c}^i$ need to be shared only between overlapping cliques. The entire algorithm can thus be run in a distributed fashion.

%% file: _sections/results1.tex
\begin{figure}[tb]
  \centering 
  \begin{minipage}{.49\linewidth}
  \centering
  \footnotesize{CT-RO}\\

 \vspace{0.4em}
 \includegraphics[width=\linewidth]{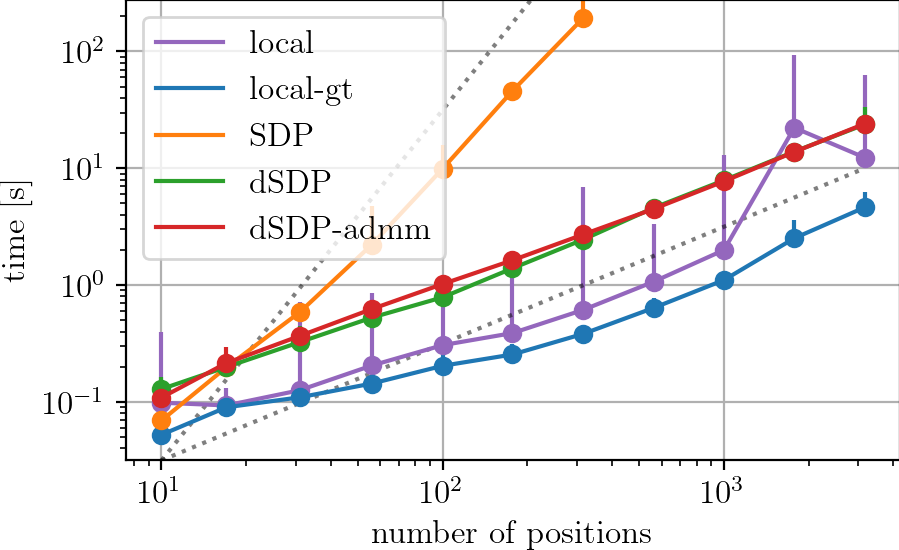}
  \end{minipage}
  \begin{minipage}{.49\linewidth}
  \centering
  \footnotesize{MW} \\

 \vspace{0.4em}
 \includegraphics[width=\linewidth]{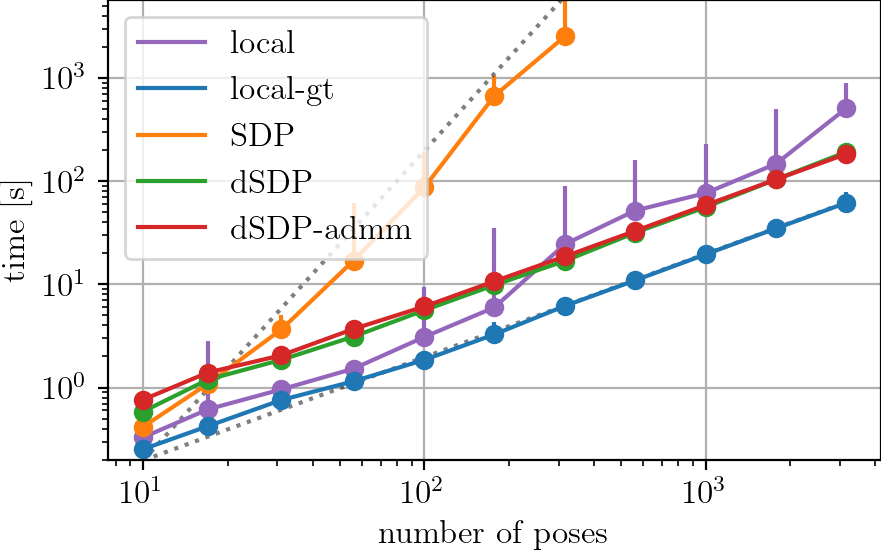}
  \end{minipage}
  \caption{Solver times for increasing problem sizes for range-only localization (left) and matrix-weighted localization (right). The dashed gray lines correspond to $O(N)$ and $O(N^3)$. The presented sparsity-exploiting algorithms have a complexity comparable with the local solver and with little variance. On average, the local solver is faster than the decomposed methods but has a high variance (vertical bars) depending on the initialization.} 
  \label{fig:timing-study}
\end{figure}

\section{Localization Results}\label{sec:results}

We test the two methods in simulation on the two example localization problems, evaluating their performance in terms of efficiency, optimality, and accuracy, compared with local and standard \ac{SDP} solvers. 

\subsection{Simulation Setup}

\vspace{1em}
\exampleheadRO We draw $N_m$ landmarks uniformly at random from a cube of dimension \SI{10}{m}. 
Then, within the same cube, we generate a random trajectory of $N$ states at times sampled uniformly at random and sorted from $0$ to $N-1$. We use the constant-velocity motion prior~\cite[Section IV.A]{barfoot_batch_2014} with random initial velocity of \SI{0.1}{ms^{-1}} and \ac{STD} $\sigma_a=\SI{0.2}{ms^{-2}}$, making sure the trajectory stays within the cube.
  Then, we simulate distance measurements by adding i.i.d Gaussian noise of \ac{STD} $\sigma_d=\SI{10}{cm}$, unless stated otherwise, to each squared distance. 
  We set the prior distance measurement noise to $\check{\sigma}_d=\SI{1}{mm}$ and $\check{\sigma}_a=\SI{0.2}{ms^{-2}}$, set $\vc{W}_{k}=\check{\sigma}_d^{-2}\vc{I}_{N_k}$ and form $\vc{W}_{ij}$ according to the~\ac{GP} prior, as explained in~\cite{barfoot_batch_2014}.

  \exampleheadMW We generate $N_m$ landmarks uniformly at random from a cube of dimension \SI{1}{m} centered at the origin. 
  Then, we sample poses uniformly at random with coordinates $\vc{t}^0_{i0}$ in $(-\SI{0.1}{m},-\SI{0.1}{m},\SI{2.9}{m})$, $(\SI{0.1}{m},\SI{0.1}{m},\SI{3.1}{m})$, with the $z$-axis pointing downwards with randomly generated yaw angles between $0$ and $2\pi$. 
We use camera parameters $f=1077$, $c=0$, $b=\SI{0.12}{mm}$ and pixel standard deviation $\sigma_u=\sigma_v=1.0$, unless stated otherwise. We generate pixel and non-isotropic $\vc{W}_k$ from these parameters by linearizing the pinhole measurement model as explained in~\cite{holmes_semidefinite_2024}. Similarly, we generate relative motion measurements and $\vc{W}_{ij}$ with translation \ac{STD} $\sigma_t=\SI{1}{cm}$ and rotation \ac{STD} $\sigma_r=\SI{10}{deg}$.

\vspace{1em}
We solve one example per problem size for the timing studies, while for the noise study, we average over ten random problem instances. For all~\ac{SDP} solvers, we use the \textit{MOSEK Fusion API for Python}~\cite{mosek_fusion}, and we set one global tolerance parameter $\tau$ per algorithm, for primal/dual feasibility, and relative duality gap parameters. For \ac{dSDP} and \ac{SDP}, we use $\tau^{\text{dSDP}}=\tau^{\text{SDP}}=10^{-10}$. For the local solvers, we use our custom \text{GN} implementations with the stopping criterion that the element-wise maximum of the gradient is below $\tau^{\text{GN}}=10^{-7}$ or at a number of iterations above $N_{\text{it}}^{\text{GN}}=100$. We consider random initializations of \ac{GN}, drawn randomly from a Gaussian of standard deviation 0.5 around the ground truth (\textit{local}), and initializing at ground truth (\textit{local-gt}).

For \altSDP, we use the penalty adaptation rule described in~\cite[eq. (3.13)]{boyd_admm_2010}, with $\mu=10$ and $\tau=2.0$. For the convergence criteria, we use thresholds on the two-norms of the primal and dual residuals as suggested in~\cite[eq. (3.12)]{boyd_admm_2010}, where we use only the relative criterion to calculate the threshold with $\epsilon^{\text{rel}}=10^{-10}$ or maximum number iterations $N_{\text{it}}^{\text{dSDP-admm}}=10$. Because \altSDP~does not require its inner iterations to be very precise, we set the tolerance $\tau^{\text{dSDP-admm}}$ to $10^{-3}$. 
We run all algorithms on a server with 2 \textit{Intel Xeon} processors, resulting in 24 physical cores and 48 threads, and with \SI{320}{GB} of RAM.

\subsection{Results}

\subsubsection{Efficiency.} First, we investigate the speed of the proposed methods as a function of the problem size. Figure~\ref{fig:timing-study} shows the performance of the two proposed algorithms as a function of the number of poses or positions. The decomposed methods are significantly faster than the full~\ac{SDP} method. In fact, these decomposed methods are competitive with the local~\ac{GN} solver (same runtime complexity), yet can still provide a certified globally optimal solution. 

Due to hardware limitations and communication overhead, the optimal number of threads for \altSDP~is fixed to 10, and is run for a fixed number of iterations $3$. We expect the performance of \altSDP, currently on par with \ac{dSDP}, to improve as more computational nodes are available. 

\begin{figure}[tb]
  \centering 
  \begin{minipage}{.49\linewidth}
  \centering
  \footnotesize{CT-RO}\\

 \vspace{0.4em}
 \includegraphics[height=3.7cm,trim={0 0 4cm 0},clip]{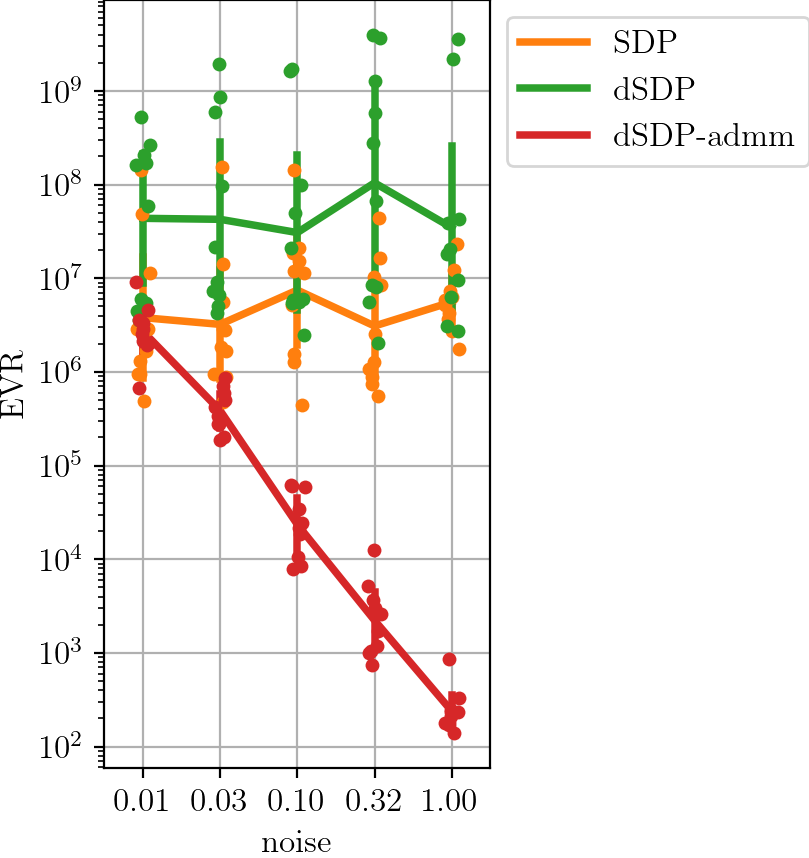}
 \includegraphics[height=3.7cm]{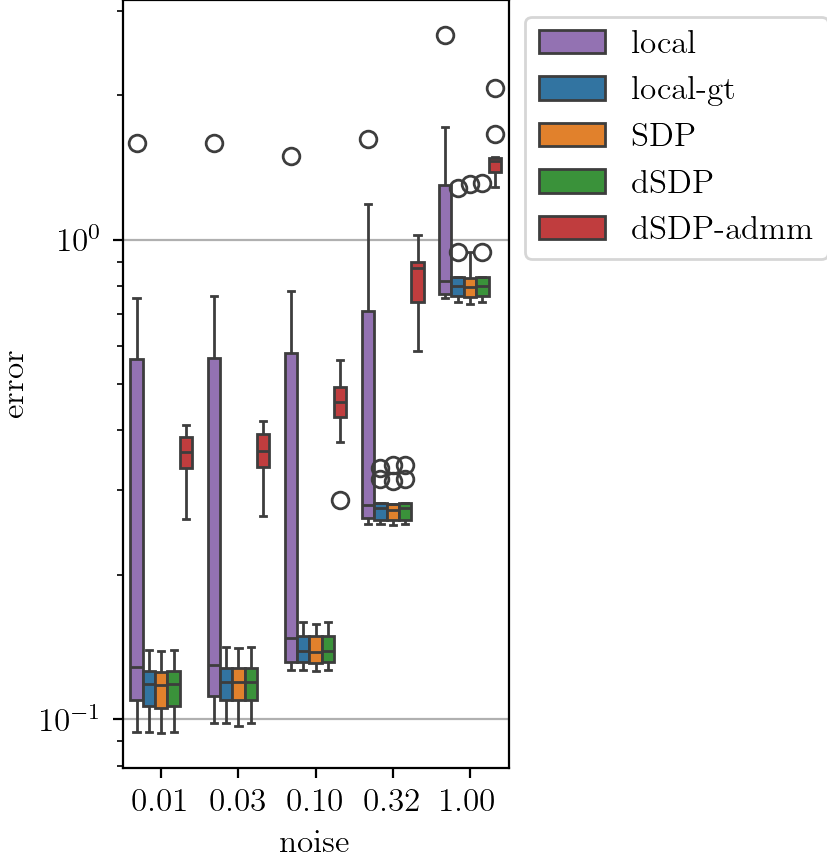}
  \end{minipage}
  \begin{minipage}{.49\linewidth}
  \centering
  \footnotesize{MW}\\

 \vspace{0.4em}
 \includegraphics[height=3.7cm,trim={0 0 4cm 0},clip]{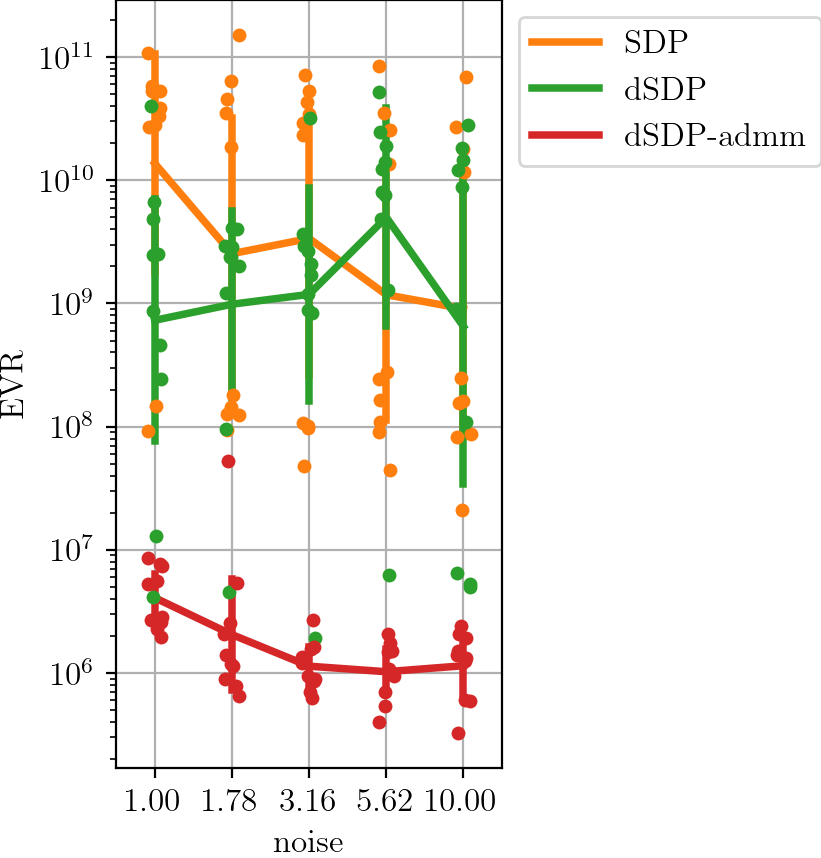}
 \includegraphics[height=3.7cm]{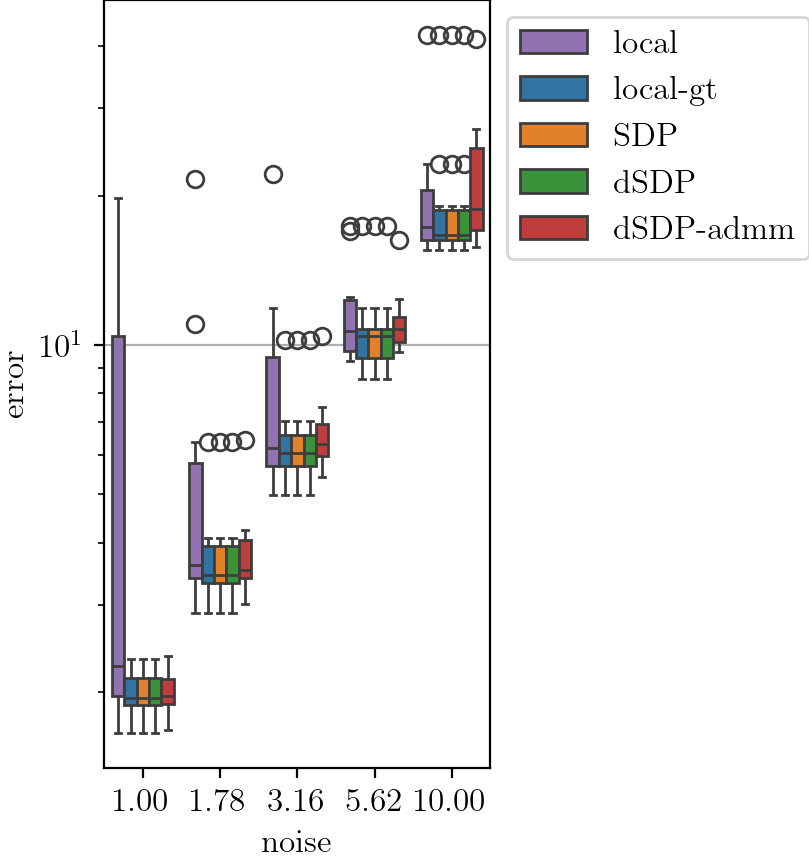}
  \end{minipage}
  \caption{Tightness and accuracy studies for increasing noise levels for range-only localization (left) and matrix-weighted localization (right). Noise is added to the squared distance measurements (in meters) and pixel observations. All results are with $N=100$ and $N_m=8$.}
  \label{fig:tightness}
\end{figure}

\subsubsection{Optimality.} Next, we investigate the tightness of our relaxations with respect to the noise level.  To this end, we compute the \ac{EVR} $\lambda_0/\lambda_1$ with $\lambda_0,\lambda_1$ the largest and second-largest eigenvalues of the solution of the relaxation $\vc{X}^*$, respectively. This ratio measures how close the solution is to being rank one.
Looking at the~\ac{EVR} in Figure~\ref{fig:tightness},
we observe that the relaxation becomes less tight with increasing noise levels, as expected. For \exampleRO~in particular, the fact that we terminate~\altSDP~after a fixed number of iterations means that the solution quality deteriorates as the noise increases. We argue that this is due to a looser connection between nodes; consisting only of a motion prior as opposed to pose-to-pose measurements. It may thus be harder for the alternating algorithm to reach `consensus'; for higher noise levels, the different subproblems pull in different directions at each iteration. Here, we see the fundamental speed \emph{vs.} accuracy trade when using the~\ac{ADMM} algorithm. We have opted for speed here, and the~\ac{ADMM} algorithm struggles to yield accurate results in the given fixed time. The \ac{SDP} and \ac{dSDP} algorithms, on the other hand, reach consistently high \ac{EVR} for all noise levels for \exampleRO. 

For \exampleMW, we observe that adding redundant constraints significantly affects the tightness of the relaxation and renders the \ac{EVR} almost constant for all considered noise levels. The added computational cost for achieving higher tightness is minimal, as shown in Figure~\ref{fig:timing-study}. We observe again that~\altSDP~is significantly less tight than the other algorithms but still achieves high enough \ac{EVR} to suggest a rank-one solution. 

\subsubsection{Accuracy.} Finally, the accuracy of the algorithms is measured in terms of the distance between ground truth values and estimates. We choose the Euclidean distance for vector-valued functions and the chordal distance for rotation estimates. The accuracy obtained by the different solvers is reported in Figure~\ref{fig:tightness}. The randomly initialized local solver (\textit{local}) has a high variance because it converges to local minima. The \textit{local-gt} solver, \ac{SDP}, and \ac{dSDP}, on the other hand, solve the problem to almost identical accuracy. Only \altSDP~has a significantly worse accuracy, particularly for the \exampleRO~example, because it would require more than the imposed three iterations to converge. This emphasizes the typical trade-off between accuracy and runtime present for \altSDP. With an improved implementation of \altSDP\footnote{The current implementation uses Python, which does not allow for effective multithreading.}, the method could be distributed more effectively and run with higher accuracy without compromising speed.  

%% file: _sections/conclusion.tex
\section{Conclusion and Future Work}\label{sec:conclusion}

We have shown that a large class of optimization problems that occur in robotics allow for linear-time~\ac{SDP} solvers using the chordal decomposition of the underlying aggregate sparsity graph. This decomposition can be used to speed up centralized computation but also lends itself to an ADMM-based distributed implementation, which we also tested. The methods are on par with efficient local solvers in terms of computational complexity, but with the important advantage that they are certifiably optimal for reasonable noise levels. 
Possible improvements include the adoption of known heuristics for speeding up our vanilla ADMM, such as diagonal rescaling for better conditioning~\cite{giselsson_diagonal_2014}, or the use of an approximate interior-point solver~\cite{rontsis_efficient_2022}. The behavior of the algorithms when the problem is not chordally sparse presents another interesting direction of research, using, for example, factor-width decomposition~\cite{zheng_chordal_2021}, loopy belief propagation~\cite{ortiz_visual_2021}, or using the chordal reduction as an initialization technique for more exact solvers.

%% file: _sections/appendix.tex
\appendix

\section{Appendix}


\subsection{General formulation}\label{app:general}
We assume that the functions $\bm{\ell}_k(\cdot)$ in \eqref{eq:xik} include enough higher-order terms in $\vc{\xi}_{i}$ so that we can write $\vc{e}_k(\vc{\xi}_k) = \vc{B}_k \vc{x}_k(\vc{\xi}_k)$ and $\vc{e}_{ij}(\vc{\xi}_i,\vc{\xi}_j) = \vc{B}_{ij}\vc{x}_{ij}(\vc{\xi}_i,\vc{\xi}_j).$
  This allows us to rewrite the cost terms in the required form by introducing 
$\vc{Q}_k:=\vc{B}_k^\top\vc{W}_k\vc{B}_k$ and $\vc{R}_{ij}:=\vc{B}_{ij}^\top\vc{W}_{ij}\vc{B}_{ij}$. We further assume that each term of the lifted vectors can be enforced with a quadratic constraint, 
which is always possible when the substitutions are polynomial or rational functions of any order.\footnote{To see this, note that for an original cost of degree $k$, we can add all monomials up to degree $k$, in which case each term can be recursively constrained using its preceding terms. In practice it is often desirable to use as few terms as possible~\cite{sparse-Lasserre}.}
Finally, assuming that the feasible sets $\mathcal{X}_n$ can also be characterized by quadratic constraints, which we denote by $\vc{A}_{i}^p$, we obtain the generic equivalent \ac{QCQP} formulation~\eqref{eq:qcqp-split}.

\subsection{Example reformulations}\label{app:examples}

\vspace{1em}
\exampleheadRO
By using just one substitution: $\ell_k(\vc{\xi}_k):=\norm{\vc{t}_k}^2$, we get the lifted vector $\vc{x}_k(\vc{\xi}_k)^\top:=\bmat{1 & \vc{t}_k^\top & \vc{v}_k^\top & \norm{\vc{t}_k}^2}$ and we can write each absolute error term as
  \begin{equation}
    \begin{aligned}
      \vc{e}_k(\vc{\xi}_k)[i] &= \bmat{\tilde{d}_{ki}^2-\norm{\vc{m}_i}^2 & 2\vc{m}_i^\top & \vc{0}_3^\top & -1}\vc{x}_k(\vc{\xi}_k)=\vc{b}_{ki}^\top \vc{x}_k(\vc{\xi}_{ki}),
  \end{aligned}
\end{equation}
where $\vc{b}_{ki}$ corresponds to one row of $\vc{B}_k$ from Section~\ref{app:general}. Similarly, we have:
  \begin{equation}
    \begin{aligned}
  \vc{e}_{ij}(\vc{\xi}_i,\vc{\xi}_j)&=\bmat{\vc{0}_6 & \vc{I}_6 & \vc{0}_6 & \vc{\Phi} & \vc{0}_6}\vc{x}_{ij}(\vc{\xi}_i,\vc{\xi}_j).
    \end{aligned}
  \end{equation}
  Since the substitution is the squared norm of the position, it is a polynomial and can thus be trivially written in quadratic form.

\vspace{1em}
\exampleheadMW
For this example, we note that the cost is already quadratic in the constraints, so no substitutions are necessary. Furthermore, the feasible set is given by $\mathcal{X}_n:=\SE{3}$, which can be enforced by the quadratic constraints $\vc{C}_{i0}^\top\vc{C}_{i0}=\vc{I}_3$ and $\vc{C}_{i0}[1] \times \vc{C}_{i0}[2] = \vc{C}_{i0}[3]$ (with $\vc{C}_{i0}[j]$ the $j$-th column of $\vc{C}_{0i}$), which enforces handedness and thus substitutes the (cubic) determinant constraint in the definition of $\SO{3}$~\cite{briales_convex_2017}.